\documentclass{article} % For LaTeX2e
\usepackage{iclr2023_conference_tinypaper,times}

\usepackage{amsmath,amsfonts,bm}

\def\eqref#1{equation~\ref{#1}}
\def\1{\bm{1}}

\DeclareMathAlphabet{\mathsfit}{\encodingdefault}{\sfdefault}{m}{sl}
\SetMathAlphabet{\mathsfit}{bold}{\encodingdefault}{\sfdefault}{bx}{n}

\usepackage{hyperref}
\usepackage{url}
\usepackage{booktabs}
\usepackage{graphicx}
\usepackage{multirow}
 \usepackage{amsmath}
 \usepackage{tabularx}
 \usepackage{lipsum}
\usepackage{xcolor}

\title{Beyond Uniform Scaling: Exploring Depth Heterogeneity in Neural Architectures}
\author{Akash Guna R.T*, Arnav Chavan* \\
Nyun AI\\
\texttt{\{akash.guna,arnav.chavan\}@nyunai.com}\\
\AND
Deepak Gupta \\
Transmute AI Lab \\
\texttt{guptadeepak2806@gmail.com} \\
}

\iclrfinalcopy % Uncomment for camera-ready version, but NOT for submission.
\begin{document}

\maketitle

\begin{abstract}
Conventional scaling of neural networks typically involves designing a base network and growing different dimensions like width, depth, etc. of the same by some predefined scaling factors. We introduce an automated scaling approach leveraging second-order loss landscape information. Our method is flexible toward skip connections, a mainstay in modern vision transformers. Our training-aware method jointly scales and trains transformers without additional training iterations. Motivated by the hypothesis that not all neurons need uniform depth complexity, our approach embraces depth heterogeneity. Extensive evaluations on DeiT-S with ImageNet100 show a 2.5\% accuracy gain and 10\% parameter efficiency improvement over conventional scaling. Scaled networks demonstrate superior performance upon training small-scale datasets from scratch. We introduce the first intact scaling mechanism for vision transformers, a step towards efficient model scaling.

% Executing large transformers is costly, and reducing parameters post-training typically requires fully retraining these models. To address this challenge, we propose a novel approach for efficiently growing the parametric space of transformers during training, thus avoiding the need for complete retraining. Unlike existing growth mechanisms that focus on scaling small transformers through incremental transfer learning, our method enables the growth of transformers while training, preventing divergence from the loss landscape. Our mechanism selects appropriate neurons to grow by leveraging second-order information, ensuring a more informed expansion. We evaluate our growing mechanism on popular transformer architectures such as DeiT, Swin, and CaiT, specifically on ImageNet and a 100-class subset of ImageNet. Additionally, we conduct a comparative analysis with existing transformer scaling mechanisms. The results indicate that our mechanism improves the accuracy of DeiT-S in classifying ImageNet-100 by 2\%, while simultaneously reducing the parametric space by 35\%. To the best of our knowledge, our work introduces the first intact growing mechanism for vision transformers, demonstrating its potential for more efficient and effective model training.
\end{abstract}

\section{Introduction}
% \par Efforts to enhance the performance of deep learning models hinge crucially on scaling network architectures. Prominent model families like ResNet, BERT, GPT-3, and ViT \citep{resnet, bert, gpt3, vit} have consistently expanded networks by adjusting depth, width, and layer dimensions, informed by empirical observations rather than a rigorous scientific foundation . A recent study by \cite{splitting} introduced an innovative width expansion approach, leveraging a computationally efficient Hessian approximation to identify neurons linked with saddle points for growth. However, this method, limited to width expansion that disrupts skip connections in linear layers, restricting its applicability to state-of-the-art transformers.
% \par Existing attempts at scaling depth like in Net2Net\citep{net2net}, uniformly increase depth throughout the network. We contend that such uniform scaling may be sub-optimal, as different network regions may require distinct scaling proportions. Notably, current approaches lack provisions for non-uniform depth scaling of neural networks.

% \par This paper breaks away from uniform network scaling and explores depth heterogeneity for non-uniform scaling of neural architectures. Our approach embraces varied depth allocations across neurons within the same layer, allowing for adaptive and efficient utilization of network resources. Through experimental comparisons, we demonstrate that our method outperforms conventional scaling methods, achieving an accuracy gain exceeding 2.5\% while utilizing 10\% fewer parameters.
Scaling of the network architectures has been a crucial aspect of pushing the performance of deep learning models.
A range of model families, such as ResNet, BERT, GPT-3, and ViT \citep{resnet, bert, gpt3, vit}, often scale networks by augmenting the depth, width, and individual layer dimensions. These adjustments are typically derived from experimental observations. The choices made in these experiments are not arbitrary but are informed by empirical insights, although they mostly lack a rigorous scientific foundation.

A robust scaling strategy, informed by insights from the loss landscape of the base network, is essential for the creation of efficient and high-performing neural networks. In a recent study by \cite{splitting}, an innovative approach was introduced to expand the width of neural networks. This method utilizes a computationally efficient Hessian approximation to identify neurons associated with saddle points for growth. Notably, the growth is limited to width, which increases layer dimensions, disrupting skip connections in linear layers and limiting their applicability to state-of-the-art transformers.

It is important to highlight that any attempt to adapt this method or traditional depth scaling approaches typically involves uniformly increasing the depth across the entire network. We hypothesize that this is a sub-optimal choice since different parts may require different scaling proportions. However, the existing approaches are not designed to perform non-uniform scaling of the network architecture.

In this paper, we push the limits beyond the uniform scaling of the network and explore depth heterogeneity for the non-uniform scaling of neural architectures. Our approach embraces varied heterogeneity of depth across neurons within the same layer to suffice the desired complexity of representation within the network. This departure from uniform depth allocation allows for a more adaptive and efficient utilization of network resources. Through experimental comparisons, we demonstrate that the presented approach generalizes better than the conventional scaling methods with an accuracy gain of more than 2.5\% while requiring 10\% fewer parameters than its counterpart. 

%% Scaling of the network architectures has been a crucial aspect for pushing the performance of deep learning models. A variety of model families \citep{resnet, bert, gpt3, vit} simply scale networks by increasing the depth, width and individual layer dimensions randomly without any scientific methodology. A solid scaling strategy which leverages information from the loss landscape of the base network can help to create efficient and high-performing networks. \citep{Liu et al} provided an initial study on growing neural networks by utilizing a computationally efficient Hessian approximation to identify neurons associated with saddle points for growth. However their method focused on creating resource constrained networks and was restricted to simpler CNN architecture trained on small scale datasets. In this work we propose a network scaling mechanism which leverages optimization saddle point information for scaling and training the network in a joint framework. Our proposed method does not increase the total number of training iterations and hence can be directly compared with conventional scaled models. Additionally, the proposed method scales the model dynamically during training by leveraging data-aware loss landscape and hence helps to get better performing models for any target dataset. 

\section{Related Works}
\par \textbf{Hessian in Deep Learning} The second order differentiation of the loss (hessian) provides some information on the loss landscape, like the curvature of loss across the parametric space\citep{curvature_loss} and presence of saddle points\citep{bengio-eig}. Second-order optimization techniques\citep{adahessian,gupta2018shampoo,second_opt_os} utilize the curvature information as momentum to stabilize optimization. Eigenvalues of a hessian reveal the presence of saddle points. When the eigenvalues of a hessian are indefinite there is a saddle point. Escaping saddle points is crucial for fast convergence and reaching minima. Plateaus with small curvature surround saddle points. Thus, the accumulation of saddle points lead to slow convergence \citep{bengio-eig}. Neural Network training behaves like a procedure to make negative eigenvalues shift towards zero, yet negative eigenvalues still exist post-training, and the amount of remaining negative eigenvalues is indirectly proportional to its architecture\citep{lecunn-eig}. Some recent works grow neural networks by splitting at neurons with saddle points that reach convergence faster with a smaller parametric space \citep{splitting_,splitting_t,splittting_s,firefly}. These works select all neurons with saddle points to grow. Growing near negative eigenvalues with low magnitude leads to minimal improvement due to closer proximity to zero. Growing the parametric space only near neurons with large negative eigenvalues could lead to faster convergence with minimal parameters, which we investigate in this paper.

\par \textbf{Growing Neural Networks} Training large networks requires high computational costs. Growing small networks into bigger variants reduces computations while maintaining accuracy. Initially \cite{net2net} grew neuron networks by creating copies of random neurons and adjusting the output for function preservation. Randomly growing neural networks could cause limited improvements to network training. \cite{splitting} selected neurons with saddle points for growth utilizing a compute-efficient Hessian approximation and added neurons by splitting an existing neuron into two offspring. \cite{splittting_s} addressed the local optimality issue by splitting neurons into offspring with positive and negative weights. \cite{splitting_t} split neurons in an energy-aware manner to grow efficient networks. Splitting neurons requires changing the symmetry of existing neurons, altering the learned functionality of the neuron is not desirable. Adding neurons without disturbing neuron weights prevents the loss of symmetry in neurons \citep{gradmax}. The current literature grows convolutional neural networks and artificial neural networks and is not directly applicable to growing transformers mainly due to skip connections. We propose a neuron growth approach that is applicable to training transformers.

\par \textbf{Scaling Transformers} Increasing the model capacity leads to better performance and is currently the direction transformer research moves towards. Increasing model capacity also increases computation requirements. Scaling transformers by knowledge transfer from smaller variants is computationally feasible. \cite{bert2bert} introduced Bert2BERT an extension of function preserving\citep{net2net} growth applied to BERT transformers. \cite{automatedscaling} uses linear transformations to grow the width and depth of pre-trained transformers. Unlike previous works, we introduce a non-uniform scaling technique to grow transformers.

\section{Method}
\par The proposed non-uniform scaling approach focuses on a training-aware strategy at the individual neuron level, aiming to expedite convergence by avoiding saddle points in the loss landscape. Neurons are selected based on a criterion involving the identification of those with the smallest negative eigenvalues, crucial for escaping saddle points. This localized scaling is performed by adding new neurons to a pseudo-layer and projecting their output back to selected neurons through individual skip connections. The method addresses challenges posed by complex network topologies, such as skip connections, by scaling the transformer with the addition of outputs from new and selected neurons while maintaining original dimensions. To preserve function during scaling, the technique incorporates the addition of neurons without modifying existing weights and biases, ensuring the preservation of intermediate functions.
\subsection{Workflow Description}
\par
\begin{figure}[h]
    \centering
    \includegraphics[width = \linewidth]{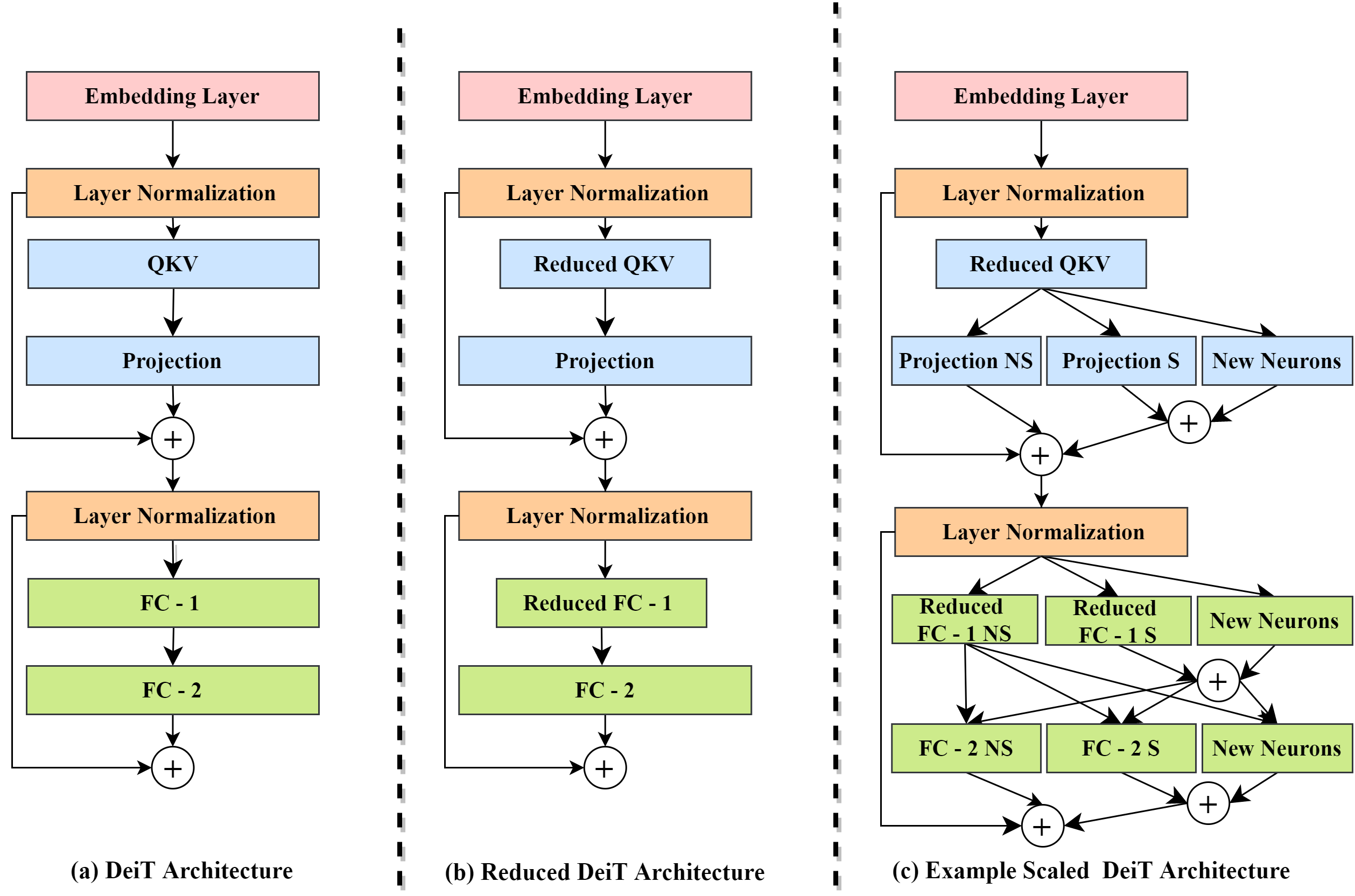}
    \caption{The Basic Workflow of the Proposed Scaling Technique. (a) shows the regular DeiT architecture. (b) shows a reduced DeiT architecture where parameters are reduced from intermediate layers to form bottlenecks. (c) shows an example of scaled DeiT architecture grown from a reduced DeiT architecture. We scale only selected neurons (S) for scaling through skip connections and do not scale other neurons (NS) present in the layer. Our scaling technique is applicable to QKV, Projection and Fully Connected layers of DeiT. }
    \label{fig:basic}
\end{figure}

 Our aim is to scale neural networks through depth heterogeneity. To achieve the scaling, we start from transformers with reduced width at intermediate layers. We reduce the width of only intermediate layers to leave skip connections unaffected. We scale neurons by adding new neurons as skip connections to neurons selected to scale. Figure \ref{fig:basic} displays the basic workflow of our scaling technique.
\subsection{Using Hessian to Identify the Right Neurons}
\par  Neural networks aim to escape saddle points to reach local minima \citep{bengio-eig,lecunn-eig}. Based on this motivation, we use the identification of saddle points as a criterion for scaling the network.  We select neurons using Hessian to identify the right neurons to scale. The second-order derivative of the activation value represented by a neuron concerning the loss is a Hessian matrix. It monitors the direction of the gradients, the first-order derivative. Therefore, we could identify saddle points upon analyzing the magnitude of the Hessian's eigenvalues. A saddle point exists when a Hessian possesses a mix of positive and negative eigenvalues. The magnitude of the negative eigenvalues provides information on the steepness of the saddle point.
\par Despite the Hessian's ability to effectively identify saddle points, they are costly to compute. \cite{splitting} utilized a compute-efficient Hessian approximation called the splitting matrix to scale neurons by splitting existing neurons. Let the loss computed over the activation value $\sigma$ of a neuron, L($\sigma$), be $\phi(\sigma(X))$ where $\phi$ is the loss concerning the $\sigma$ for the given input $X$. The Hessian for  a $\sigma$ is given by
\begin{equation}
    \label{eq:hess}
    Hessian(\sigma) = \phi^{''}(\sigma(X)) \sigma^{'}(X)^T  \sigma^{'}(X)  + \phi^{'}(\sigma(X))  \sigma^{''}(X)
    \end{equation}
where $(.)^{'}$ represents the derivative of first order and $(.)^{''}$ represents the second order derivative. Calculating $\phi^{''}$ for a $\sigma_i$ requires computing all the preceding gradients $\{\sigma^{'}_n, \sigma^{'}_{n-1},...,\sigma^{'}_{i-1}\}$ which constitutes the bulk of computation required for Hessian calculation. Omitting $ \phi^{''}(\sigma(X)) \sigma^{'}(X)^T  \sigma^{'}(X)$ from computation was still able to produce a good enough approximation \citep{splitting} and is known as splitting matrix. The splitting matrix is formally defined as
    \begin{equation}
    \label{eq:split}
        Splitting Matrix( \sigma) = \phi^{'}(\sigma(X))  \sigma^{''}(X)
    \end{equation}
\par We use the eigenvalues from the splitting matrix to identify neurons with saddle points that are suitable to scale.
\subsection{Selecting Neurons to Scale}
\par Our base network is a DeiT-S with reduced parameters, which we scale to 20 million parameters during training. We train that network for an \textit{Intial Warmup} period of 50 epochs. Starting from the 50th epoch, we scale the network every 30 epochs. We denote this interval as the \textit{Scaling Interval}. Neurons with the smallest negative eigenvalues are selected for scaling during each splitting interval. 
\par We select the neurons until the added neurons exhaust a \textit{Parameter Budget}. Parameter Budget is defined as the expected count of newly added neurons at each scaling interval. Since we are reducing the parameters from both MHSA and MLP blocks we scale neurons present in both these blocks. The resulting scaled network adds parameters to both MLP and MHSA blocks. We use a \textit{Layer Threshold} to set a minimum bar for the number of eligible neurons a layer must possess to be scaled. This ensures that sparse neuron scaling is prevented. We set the layer threshold to 60 neurons for scaling a Deit-S transformer.

\subsection{Scaling neurons}
\par We scale transformers at the neuron level. Due to extensive usage of skip connections between transformer blocks, existing neuron level scalers like \cite{net2net,splitting} are not directly applicable. We overcome this issue by adding neurons in a manner that mimics skip connections. We add outputs of added neurons with their existing counterparts without altering the existing output dimensional space of the transformer blocks. Our architecture initializes added neurons such that existing neuron outputs and weights remain unchanged, ensuring function preservation \citep{net2net}.
\par Since the widely adopted transformer architecture \citep{transformer} is constructed using linear layers, we design our scaling technique around linear layers. Let the output of a layer that consists of neurons selected to be scaled be $O_L(I) = W_L \times I + B_L $ for a given input $I$ where $W_L$ and $B_L$ are the weights and bias of the layer respectively. Our goal is to expand a subset of neurons $N_S$ from a layer consisting of $N_L$ neurons without disturbing the weights$W_S$ and bias $B_S$. To achieve this, we perform scaling by adding two neurons for each selected neuron with equal weights $W_{A}$ and biases $B_{A}$ but opposite polarities.  $W_{A}$ and $B_{A}$ are $W_{S}$ and $B_{S}$ scaled by a \textit{scaling factor} of 0.2. This initialization helps in preserving $W_S$ and $B_S$ while ensuring the flow of gradients. The equation illustrating how our scaling mechanism by adding a skip connection between new neuron outputs and existing outputs of selected neurons is given by:
\begin{equation}
    \label{eq:splitting_mechanism}
    O^{'}_{S} = O_{S} + \underbrace{GeLU( O_{A+} + O_{A-} ) + O_{A+} + O_{A-}}_{Neuron Scaling}
\end{equation}
where $O_S$ and $O^{'}_{S}$ are outputs of a neuron from $N_S$ before and after scaling. Further, GeLU(.) adds non-linearity to the outputs of newly added neurons $(O_{A+}, O_{A_-})$. When initialized, the addition between $O_{A+}$ and $O_{A-}$ becomes 0 leaving $W_S$ and $B_S$ unmodified. We have skip connections from $O_{A+}$ and $O_{A_-}$ to ensure proper gradient flow during backpropagation when newly initialized.

\section{Training Details}
\subsection{Dataset Description}
\par We utilize a 100-class subset of ImageNet-1000 \citep{imagenet} for experimentation. ImageNet-1000 was created for the ImageNet Large Scale Visual Recognition Challenge (ILSVRC), a competition that spurred advancements in image classification algorithms. ImageNet-1000 covers many object categories, including animals, vehicles, household items, and more. Neural Networks that learn to classify ImageNet tend to learn generic feature descriptors; therefore, they are often chosen as base networks for transfer learning. We also use CIFAR-100 \citep{cifar}, a medium-scale dataset containing 100 categories, to show that the scaled network performed better than Deit-S at learning medium-scale datasets from scratch. 
\subsection{Hyperparameters}
\par For all experiments using ImageNet-100 we utilize a batch size of 1024 equally distributed among 4 NVIDIA L4 GPUs. We train all networks for 300 epochs and adopt the same training hyper-parameters as in \cite{deit} unless specified explicitly. For the CIFAR-100 experiments we set batch size to 512 (/4 GPUs), gradient clipping to 1.0, and weight decay to 0.0001. 

\subsection{Reducing Memory utilization using  shared input tensors}
\par During forward propagation, each linear layer in PyTorch retains an individual copy of the input to compute gradients during backward propagation. This mitigates the risk of erroneous gradient calculations resulting from input modification during forward propagation. When multiple layers share the same input, pytorch stores redundant copies of the input increasing memory usage. When we grow transformers, we find that accumulation of redundant copies exponentially increases memory utilization. To prevent such accumulation, we used a hash table to store a single shared input copy between all layers using the input which led to memory-efficient gradient calculation. The key of the hash table is the 32-bit mean of the input. The hash table used chaining to store different inputs with the same mean. 
{
\section{Experiments}
\par We conduct experiments to analyze the performance of scaled DeiT-S in various settings. We also identify appropriate reduction and scaling hyper-parameters. Finally, we show that our scaling techniques aid transformers in escaping saddle points during transformer training.
\subsection{Comparison with unscaled DeiT-S}
\par Here we show the performance of the scaled network $(\text{Reduced DeiT-S} \rightarrow \text{DeiT-S})$ compared to unscaled DeiT-S on classifying ImageNet-100 dataset. To scale DeiT-S, we first trimmed the width of intermediate fully connected layers. The reduced network contained 11.0 M parameters, a 10.7 M parameter decrease from the original DeiT-S with 21.7 M. The reduced DeiT-S outperformed DeiT with 15.6 M parameters (28\% lesser parameters) and received over 2.5 \% increase in accuracy with 19.4 M parameters over DeiT-S, that is, a 10 \% reduction in parameters. Table \ref{tab:imnet100} displays the experiment results.

\begin{table}[htbp]
    \centering
    \resizebox{\textwidth}{!}{
    \begin{tabular}{lcccccc}
        \toprule
        Scaling & Base Param. (M) & Final Param. (M) & Base FLOPs (G) & Final FLOPs (G) & Top-1 & Top-5 \\
        \midrule
        Homogeneous & 21.7  & 21.7  &4.6 & 4.6& 77.80 & 93.16 \\
        Heterogeneous  & 11.0 & 15.6  & 2.3& 3.1& 79.16 & 94.00 \\ 
        Heterogeneous & 11.0  & 19.4  & 2.3 & 3.9& 80.36 & 94.58 \\ 
        % Deit-T & & & & & & \\
        % Deit-T & 2.9  & 4.5  & & & & \\
        % Deit-T & 2.9  & 5.9  & & & 74.20 & 91.74 \\
        \bottomrule
    \end{tabular}}
    \caption{Performance of the proposed scaling method to scale DeiT-S on ImageNet100 dataset.}
    \label{tab:imnet100}
\end{table}

\subsection{Training CIFAR-100 from Scratch}
\par Using the scaled DeiT-S network on Imagenet-100, we train CIFAR-100, a medium-scale dataset, from scratch. This experiment identifies that our scaled network overcame the inability of handmade transformers to achieve good performance upon training small datasets from scratch \citep{vit,deit}. We hypnotize that the DeiT-S underperforms in CIFAR-100 classification due to fast overfitting of data which our scaled network averts. Table \ref{tab:mytable2} shows the performance of scaled DeiT-S and unscaled Deit-S for training CIFAR-100 from scratch.
% \begin{figure}[htbp]
%     \centering
%     \begin{minipage}{\textwidth}
%         \centering
%         \resizebox{0.6\textwidth}{!}{
%         \begin{tabular}{ccccc}
%             \hline
%             Model & Param. (M) & FLOPs (G) & CIFAR10 & CIFAR100 \\
%             \hline
%             Deit-S (No Growth) &21.7M  & 4.6 & 58.9 & 9\\
%             Deit-S (Grown) & 19.4M  & 3.9  &  95& 78.0\\
%             \hline
%         \end{tabular}}
%         \caption{Your table caption here.}
%         \label{tab:mytable4}
%     \end{minipage}%
% \end{figure}

\begin{table}[htbp]
\centering
        \begin{tabular}{lcccc}
            \hline
            Model & Param. (M) & FLOPs (G) & Top-1 & Top-5 \\
            \hline
            Deit-S (Homogeneous) &21.7M  & 4.6 & 58.9 & 78.9\\
            Deit-S (Heterogeneous) & 19.4M  & 3.9  &  78.1& 95.0\\
            \hline
        \end{tabular}
\caption{Performance on our scaled DeiT-S on CIFAR100 upon training from scratch.}
\end{table}
\subsection{Identifying the ideal base model}
\par Since we scale DeiT transformers with reduced parameters, it is important to ideally reduce parameters to maximize the effectiveness of scaling. We reduce parameters from the base network by reducing the width of intermediate layers, forming bottlenecks. We reduce parameters from QKV layers (ATTN) and the first fully connected layer of the MLP block (FC). Other layers have skip connections therefore, width cannot be modified. We performed a grid search to identify the best ratio at which parameters could be reduced from ATTN and FC layers. We tried reducing the width of ATTN and FC layers at equal proportions and reducing the width of only ATTN or FC Layers.  Table \ref{tab:mytable2} shows the result of the grid search. Reducing the width by half at both ATTN and FC layers produced the best results. 

\begin{table}[htbp]
    \centering
    \resizebox{\textwidth}{!}{
    \begin{tabular}{lcccccccc}
        \toprule
        Model & Base Param. (M) & Grown Param. (M) & Base FLOPs (G) & Grown FLOPs (G) & FC $\downarrow$ & ATTN $\downarrow$ & Top-1 & Top-5 \\
        \midrule
        Deit-S & 11.0  & 19.4  &2.3 & 3.9& /2 & /2 & 80.36 & 94.58 \\
        Deit-S & 11.4  & 19.7  & 2.4 & 4.1& /4 & /1 & 78.64 & 93.88 \\
        Deit-S & 17.0 & 19.4 & 3.4 &3.8& /1 & /4 & 79.44 & 93.36 \\
        Deit-S & 21.7  & 21.7  & 4.6 & 4.6& /1 & /1 & 77.80 & 93.16 \\
        \bottomrule
    \end{tabular}}
    \caption{Effect of base network reduction on final performance on the ImageNet100 dataset.}
    \label{tab:mytable2}
\end{table}

\subsection{Identifying the appropriate scaling interval}
\par To explore the ideal splitting interval, we searched across a range of candidate intervals \{10,20,30,50\} to scale a reduced Deit-S transformer. We scale the DeiT-S transformer by reducing intermediate layers in MLP and MLHA blocks by 50 \% in a 1:1 ratio, resulting in a base network with 11 million parameters. We then grew all the models to 20 million parameters with an $\pm .5$ million allowed fluctuation in the final parameter count. Our experimental investigation revealed that scaling DeiT-S at a regular interval of 30 epochs yielded optimal performance while possessing the least amount of parameters. Table \ref{tab:mytable3} shows the results of the experimentation. 
\begin{table}[htbp]
    \centering
    \begin{tabular}{cccccc}
        \hline
        Grown Param. (M) & Grown FLOPs (G) & Splitting Interval  & Top-1 & Top-5 \\
        \hline
        20.4& 4.2 & 10 &79.78 &94.36 \\
        20.0&4.1 & 20 &79.44& 94.04\\
        \textbf{19.4}&3.9 & \textbf{30} & \textbf{80.36} & \textbf{94.58}\\
        19.9&4.1 & 50 & 79.94 & 93.92 \\
        \hline
    \end{tabular}
    \caption{Results on ablating scaling interval for a DeiT-S transformer trained on ImageNet100 dataset. Scaling Interval is the interval between two  scalings.  }
    \label{tab:mytable3}
\end{table}

\subsection{Analysis of Eigenvalues}
\par Examining the presence of saddle points at the start and end of training helps assess the effectiveness of our scaling method. We plot negative eigenvalues to assess the presence of saddle points and their steepness. Negative eigenvalue plots for a QKV layer from the MHSA block and the first FC layer(FC1) from the MLP block are shown in Figure \ref{fig:neg_plot}. The plots show the presence of saddle points at both the start and end of the training, but the magnitude of negative eigenvalues is very close to 0 towards the end of the training, denoting that the remaining saddle points are shallow. These plots show that the scaled transformer successfully escaped steep saddle points.
\begin{figure}[h]
    \centering
    \includegraphics[width=\linewidth]{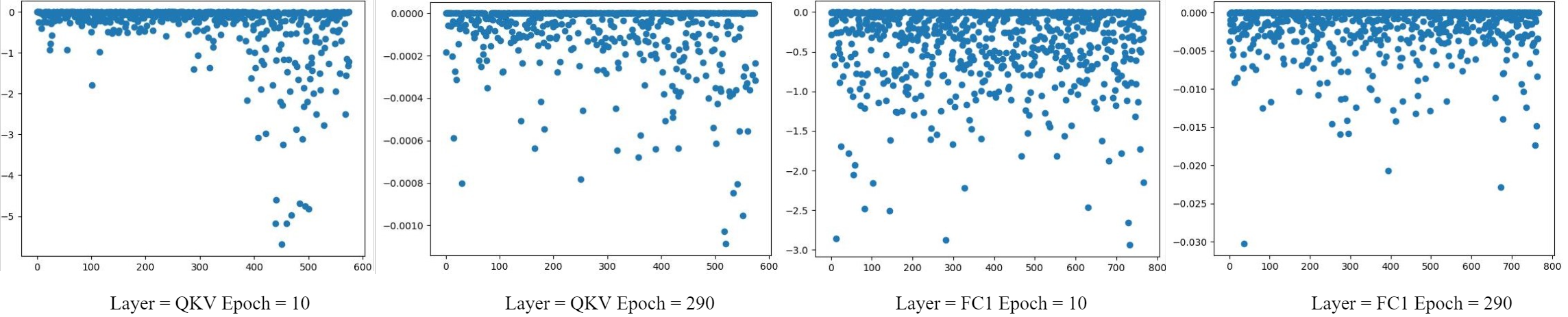}
    \caption{Plots showing negative eigenvalues of neurons in QKV and FC1 layers of the first transformer block. Each neuron in the X-axis has its magnitude shown in the Y-axis. (Zoom to view X and Y axes). }
    \label{fig:neg_plot}
\end{figure}

\subsection{Function preservation upon Initialization}
\par Here, we present proof that our scaling mechanism preserves functions during initialization from a layer perspective. Let us denote newly added positive and negative neurons collectively as linear layers ($L_{A+}$ and $L_{A-}$). $L_{A+}$ and $L_{A-}$ differ by polarity of their weights and biases. The outputs of $L_{A+}$ and $L_{A-}$ is denoted as $O_{A+}$ and $O_{A-}$ respectively and are defined as
\begin{equation}
    \label{eq:new_layer_def}
    O_{A\pm} =(\pm W_S)  I + (\pm B_S)
\end{equation}
Let {S} denote the sum of $O_{A+}$ and $O_{A-}$
\begin{align}
    &S = O_{A+} + O_{A-} \\
    &S =  W_S  I + B_S - W_S  I - B_S \\
    &S = W_S ( I - I)  + B_S (Id- Id) \\
    &S = W_S  (Z^{I}) + B_S(Z^{Id}) \\
    &S = Z^{S}
\end{align}
where $Z^{\alpha}$ is a null matrix of $\alpha$ dimensions and $Id$ is the identity matrix. Therefore, $S$ becomes equal to a null matrix of the same dimensionality. When we substitute $S$ in Equation \ref{eq:splitting_mechanism} we get 
\begin{align}
     &O^{'}_{S} = O_{S} + GeLU(S) + S \\
     &O^{'}_{S} = O_{S} + GeLU(Z^{S}) + Z^{S} 
\end{align}
 Since $GeLU(Z^{S}) = Z^S$, $O^{'}_{S} = O_{S}$. Therefore, we have shown that functions are preserved upon initialization.
\section{Conclusion}
\par In this paper, we presented a novel neural architecture scaling approach to scale modern neural architectures efficiently through localized and non-uniform scaling of neurons. We showed that the scaled DeiT-S outperforms the original DeiT-S by a margin of over 2.5 \% with 10 \% lesser parameters. We have also shown that the scaled DeiT-S is superior to its unscaled counterpart for classifying CIFAR-100 (medium-scale dataset) from scratch. Our scaling strategy aided in escaping saddle points during neural network training. This work will pave the way for future research toward building large-scale, efficient neural architectures through the proposed scaling of smaller networks.
% \par In this paper, we presented a novel neural architecture scaling approach to scale modern neural architectures efficiently through localized and non-uniform scaling of neurons. Through multiple experiments, we have demonstrated the efficacy of our approach, and we believe this work will pave way for future research towards building large-scale efficient neural architectures through proposed scaling of smaller networks.

% \subsubsection*{URM Statement}
% The authors acknowledge that the first author of this work meets the URM criteria of ICLR 2024 Tiny Papers Track.
\pagebreak
\bibliography{iclr2023_conference_tinypaper}

\begin{thebibliography}{22}
\providecommand{\natexlab}[1]{#1}
\providecommand{\url}[1]{\texttt{#1}}
\expandafter\ifx\csname urlstyle\endcsname\relax
  \providecommand{\doi}[1]{doi: #1}\else
  \providecommand{\doi}{doi: \begingroup \urlstyle{rm}\Url}\fi

\bibitem[Brown~et al.(2020)]{gpt3}
Tom Brown~et al.
\newblock Language models are few-shot learners.
\newblock In H.~Larochelle, M.~Ranzato, R.~Hadsell, M.F. Balcan, and H.~Lin (eds.), \emph{Advances in Neural Information Processing Systems}, volume~33, pp.\  1877--1901. Curran Associates, Inc., 2020.

\bibitem[Chen et~al.(2022)Chen, Yin, Shang, Jiang, Qin, Wang, Wang, Chen, Liu, and Liu]{bert2bert}
Cheng Chen, Yichun Yin, Lifeng Shang, Xin Jiang, Yujia Qin, Fengyu Wang, Zhi Wang, Xiao Chen, Zhiyuan Liu, and Qun Liu.
\newblock bert2bert: Towards reusable pretrained language models.
\newblock In \emph{Proceedings of the 60th Annual Meeting of the Association for Computational Linguistics (Volume 1: Long Papers)}, pp.\  2134--2148, 2022.

\bibitem[Chen et~al.(2015)Chen, Goodfellow, and Shlens]{net2net}
Tianqi Chen, Ian Goodfellow, and Jonathon Shlens.
\newblock Net2net: Accelerating learning via knowledge transfer.
\newblock \emph{arXiv preprint arXiv:1511.05641}, 2015.

\bibitem[Dauphin et~al.(2014)Dauphin, Pascanu, Gulcehre, Cho, Ganguli, and Bengio]{bengio-eig}
Yann~N Dauphin, Razvan Pascanu, Caglar Gulcehre, Kyunghyun Cho, Surya Ganguli, and Yoshua Bengio.
\newblock Identifying and attacking the saddle point problem in high-dimensional non-convex optimization.
\newblock \emph{Advances in neural information processing systems}, 27, 2014.

\bibitem[Deng et~al.(2009)Deng, Dong, Socher, Li, Li, and Fei-Fei]{imagenet}
Jia Deng, Wei Dong, Richard Socher, Li-Jia Li, Kai Li, and Li~Fei-Fei.
\newblock Imagenet: A large-scale hierarchical image database.
\newblock In \emph{2009 IEEE conference on computer vision and pattern recognition}, pp.\  248--255. Ieee, 2009.

\bibitem[Devlin et~al.(2019)Devlin, Chang, Lee, and Toutanova]{bert}
Jacob Devlin, Ming-Wei Chang, Kenton Lee, and Kristina Toutanova.
\newblock {BERT}: Pre-training of deep bidirectional transformers for language understanding.
\newblock In \emph{Proceedings of the 2019 Conference of the North {A}merican Chapter of the Association for Computational Linguistics: Human Language Technologies, Volume 1 (Long and Short Papers)}, pp.\  4171--4186, Minneapolis, Minnesota, June 2019. Association for Computational Linguistics.
\newblock \doi{10.18653/v1/N19-1423}.

\bibitem[Evci et~al.(2022)Evci, van Merrienboer, Unterthiner, Vladymyrov, and Pedregosa]{gradmax}
Utku Evci, Bart van Merrienboer, Thomas Unterthiner, Max Vladymyrov, and Fabian Pedregosa.
\newblock Gradmax: Growing neural networks using gradient information.
\newblock \emph{arXiv preprint arXiv:2201.05125}, 2022.

\bibitem[Gilmer et~al.(2022)Gilmer, Ghorbani, Garg, Kudugunta, Neyshabur, Cardoze, Dahl, Nado, and Firat]{curvature_loss}
Justin Gilmer, Behrooz Ghorbani, Ankush Garg, Sneha~Reddy Kudugunta, Behnam Neyshabur, David Cardoze, George~Edward Dahl, Zachary Nado, and Orhan Firat.
\newblock A loss curvature perspective on training instability in deep learning.
\newblock 2022.
\newblock URL \url{https://arxiv.org/abs/2110.04369}.

\bibitem[Gupta et~al.(2018)Gupta, Koren, and Singer]{gupta2018shampoo}
Vineet Gupta, Tomer Koren, and Yoram Singer.
\newblock Shampoo: Preconditioned stochastic tensor optimization.
\newblock In \emph{International Conference on Machine Learning}, pp.\  1842--1850. PMLR, 2018.

\bibitem[He et~al.(2016)He, Zhang, Ren, and Sun]{resnet}
Kaiming He, Xiangyu Zhang, Shaoqing Ren, and Jian Sun.
\newblock Deep residual learning for image recognition.
\newblock In \emph{Proceedings of the IEEE conference on computer vision and pattern recognition}, pp.\  770--778, 2016.

\bibitem[Kolesnikov et~al.(2021)Kolesnikov, Dosovitskiy, Weissenborn, Heigold, Uszkoreit, Beyer, Minderer, Dehghani, Houlsby, Gelly, Unterthiner, and Zhai]{vit}
Alexander Kolesnikov, Alexey Dosovitskiy, Dirk Weissenborn, Georg Heigold, Jakob Uszkoreit, Lucas Beyer, Matthias Minderer, Mostafa Dehghani, Neil Houlsby, Sylvain Gelly, Thomas Unterthiner, and Xiaohua Zhai.
\newblock An image is worth 16x16 words: Transformers for image recognition at scale.
\newblock 2021.

\bibitem[Krizhevsky et~al.(2009)Krizhevsky, Hinton, et~al.]{cifar}
Alex Krizhevsky, Geoffrey Hinton, et~al.
\newblock Learning multiple layers of features from tiny images.
\newblock 2009.

\bibitem[Osawa et~al.(2019)Osawa, Tsuji, Ueno, Naruse, Yokota, and Matsuoka]{second_opt_os}
Kazuki Osawa, Yohei Tsuji, Yuichiro Ueno, Akira Naruse, Rio Yokota, and Satoshi Matsuoka.
\newblock Large-scale distributed second-order optimization using kronecker-factored approximate curvature for deep convolutional neural networks.
\newblock pp.\  12359--12367, 2019.

\bibitem[Sagun et~al.(2017)Sagun, Bottou, and LeCun]{lecunn-eig}
Levent Sagun, Leon Bottou, and Yann LeCun.
\newblock Eigenvalues of the hessian in deep learning: Singularity and beyond, 2017.

\bibitem[Touvron et~al.(2021)Touvron, Cord, Douze, Massa, Sablayrolles, and Jegou]{deit}
Hugo Touvron, Matthieu Cord, Matthijs Douze, Francisco Massa, Alexandre Sablayrolles, and Herve Jegou.
\newblock Training data-efficient image transformers \& distillation through attention.
\newblock In Marina Meila and Tong Zhang (eds.), \emph{Proceedings of the 38th International Conference on Machine Learning}, volume 139 of \emph{Proceedings of Machine Learning Research}, pp.\  10347--10357. PMLR, 18--24 Jul 2021.

\bibitem[Vaswani et~al.(2017)Vaswani, Shazeer, Parmar, Uszkoreit, Jones, Gomez, Kaiser, and Polosukhin]{transformer}
Ashish Vaswani, Noam Shazeer, Niki Parmar, Jakob Uszkoreit, Llion Jones, Aidan~N Gomez, {\L}ukasz Kaiser, and Illia Polosukhin.
\newblock Attention is all you need.
\newblock \emph{Advances in neural information processing systems}, 30, 2017.

\bibitem[Wang et~al.(2019)Wang, Li, Wu, Chandra, and Liu]{splitting_t}
Dilin Wang, Meng Li, Lemeng Wu, Vikas Chandra, and Qiang Liu.
\newblock Energy-aware neural architecture optimization with fast splitting steepest descent.
\newblock \emph{arXiv preprint arXiv:1910.03103}, 2019.

\bibitem[Wang et~al.(2023)Wang, Panda, Hennigen, Greengard, Karlinsky, Feris, Cox, Wang, and Kim]{automatedscaling}
Peihao Wang, Rameswar Panda, Lucas~Torroba Hennigen, Philip Greengard, Leonid Karlinsky, Rogerio Feris, David~Daniel Cox, Zhangyang Wang, and Yoon Kim.
\newblock Learning to grow pretrained models for efficient transformer training.
\newblock In \emph{The Eleventh International Conference on Learning Representations}, 2023.
\newblock URL \url{https://openreview.net/forum?id=cDYRS5iZ16f}.

\bibitem[Wu et~al.(2019)Wu, Wang, and Liu]{splitting}
Lemeng Wu, Dilin Wang, and Qiang Liu.
\newblock Splitting steepest descent for growing neural architectures.
\newblock \emph{Advances in neural information processing systems}, 32, 2019.

\bibitem[Wu et~al.(2020{\natexlab{a}})Wu, Liu, Stone, and Liu]{firefly}
Lemeng Wu, Bo~Liu, Peter Stone, and Qiang Liu.
\newblock Firefly neural architecture descent: a general approach for growing neural networks.
\newblock 33:\penalty0 22373--22383, 2020{\natexlab{a}}.

\bibitem[Wu et~al.(2020{\natexlab{b}})Wu, Ye, Lei, Lee, and Liu]{splittting_s}
Lemeng Wu, Mao Ye, Qi~Lei, Jason~D Lee, and Qiang Liu.
\newblock Steepest descent neural architecture optimization: Escaping local optimum with signed neural splitting.
\newblock \emph{arXiv preprint arXiv:2003.10392}, 2020{\natexlab{b}}.

\bibitem[Yao et~al.(2021)Yao, Gholami, Shen, Mustafa, Keutzer, and Mahoney]{adahessian}
Zhewei Yao, Amir Gholami, Sheng Shen, Mustafa Mustafa, Kurt Keutzer, and Michael Mahoney.
\newblock Adahessian: An adaptive second order optimizer for machine learning.
\newblock 2021.

\end{thebibliography}
\bibliographystyle{iclr2023_conference_tinypaper}

\end{document}